\documentclass[sigconf]{acmart}

\makeatletter
\def\@ACM@checkaffil{% Only warnings
    \if@ACM@instpresent\else
    \ClassWarningNoLine{\@classname}{No institution present for an affiliation}%
    \fi
    \if@ACM@citypresent\else
    \ClassWarningNoLine{\@classname}{No city present for an affiliation}%
    \fi
    \if@ACM@countrypresent\else
        \ClassWarningNoLine{\@classname}{No country present for an affiliation}%
    \fi
}
\makeatother

\usepackage[english]{babel}
\usepackage{balance}
\usepackage{multicol}
\usepackage[section]{placeins}
\usepackage{booktabs} % For formal tables
\usepackage{pifont}
\usepackage{adjustbox}
\usepackage{microtype} % must be loaded after all fonts
\usepackage{amsmath}
\usepackage{cleveref}
\usepackage[shortlabels]{enumitem}

\usepackage{subfigure}
\usepackage{color}
\usepackage{graphicx}
\usepackage{xspace}
\usepackage{multirow}
\usepackage{makecell} % Not sure if we need this
\usepackage[ruled,vlined]{algorithm2e}
\usepackage{float}
\usepackage{ulem}
\ifdefined\isFinalized
\newcommand{\NewCommentType}[3]{}
\else
\newcommand{\NewCommentType}[3]{\expandafter\newcommand\csname #1\endcsname[1]{{\color{#2}{#3: ##1}} }}
\fi

\copyrightyear{2023}
\acmYear{2023}
\setcopyright{licensedusgovmixed}
\acmConference[KDD '23]{Proceedings of the 29th ACM SIGKDD Conference on Knowledge Discovery and Data Mining}{August 6--10, 2023}{Long Beach, CA, USA}
\acmBooktitle{Proceedings of the 29th ACM SIGKDD Conference on Knowledge Discovery and Data Mining (KDD '23), August 6--10, 2023, Long Beach, CA, USA}
\acmPrice{15.00}
\acmDOI{10.1145/3580305.3599775}
\acmISBN{979-8-4007-0103-0/23/08}



\usepackage[override]{cmtt} % make tt font tighter / less ugly

\begin{document}

\crefformat{section}{\S#2\color{blue}#1#3} % see manual of cleveref, section 8.2.1
\crefformat{subsection}{\S#2\color{blue}#1#3}
\crefformat{subsubsection}{\S#2#1#3}

\renewcommand\cellalign{cl}
\renewcommand\cellgape{\Gape[1pt]}

\renewcommand\footnotetextcopyrightpermission[1]{} % removes footnote with conference information in first column
\providecommand{\myparab}[1]{\smallskip\noindent\textbf{#1} }
\newcommand{\sectionref}[1]{$\S$\ref{#1}}
\newcommand{\eg}{e.g.,}
\newcolumntype{?}{!{\vrule width 1pt}}
\title[Augmenting Rule-based DNS Censorship Detection at Scale with Machine Learning]{Augmenting Rule-based DNS Censorship Detection\\at Scale with Machine Learning}

\author{Jacob Brown}\authornote{Co-first authors contributed equally to this work.}
\affiliation{%
  \institution{Princeton University}
}

\author{Xi Jiang}\authornotemark[1]
\affiliation{%
  \institution{University of Chicago}
}

\author{Van Tran}\authornotemark[1]
\affiliation{%
  \institution{University of Chicago}
}

\author{Arjun Nitin Bhagoji}
\affiliation{%
  \institution{University of Chicago}
}

\author{Nguyen Phong Hoang}
\affiliation{%
  \institution{University of Chicago}
}

\author{Nick Feamster}
\affiliation{%
  \institution{University of Chicago}
}

\author{Prateek Mittal}
\affiliation{%
  \institution{Princeton University}
}

\author{Vinod Yegneswaran}
\affiliation{%
  \institution{SRI International}
}

\renewcommand{\shortauthors}{J. Brown et al.}

\begin{sloppypar}
\begin{abstract}

The proliferation of global censorship has led to the development of a plethora
of measurement platforms to monitor and expose it. Censorship of the domain name
system (DNS) is a key mechanism used across different countries. It is currently
detected by applying heuristics to samples of DNS queries and responses (probes)
for specific destinations. These heuristics, however, are both platform-specific
and have been found to be brittle when censors change their blocking behavior,
necessitating a more reliable automated process for detecting censorship.

In this paper, we explore how machine learning (ML) models can (1) help
streamline the detection process, (2) improve the potential of using large-scale
datasets for censorship detection, and (3) discover new censorship instances and
blocking signatures missed by existing heuristic methods. Our study shows that
supervised models, trained using expert-derived labels on instances of known
anomalies and possible censorship, can learn the detection heuristics employed
by different measurement platforms. More crucially, we find that unsupervised
models, trained solely on uncensored instances, can \emph{identify new instances
and variations of censorship missed by existing heuristics}. Moreover, both
methods demonstrate the capability to uncover a substantial number of new DNS
blocking signatures, i.e., injected fake IP addresses overlooked by existing
heuristics. These results are underpinned by an important methodological
finding: comparing the outputs of models trained using the same probes but with
labels arising from independent processes allows us to more reliably detect
cases of censorship in the absence of ground-truth labels of censorship.

\end{abstract}

\maketitle

\section{Introduction}
\label{sec:intro}

As the Internet becomes an indispensable medium for modern communications,
authoritarian governments are increasingly making use of technologies to control
the free flow of online information, especially in repressive
regimes~\cite{deibert2020reset, Aryan:2013, Ramesh2020DecentralizedCA,
USESEC21:GFWatch}. In response, there have been tremendous efforts by
researchers and other stakeholders committed to Internet freedom to detect and
circumvent Internet censorship around the world~\cite{Tor, filasto2012ooni,
Nobori2014, Iris, Bock2019GenevaEC, niaki2020iclab, Raman2020CensoredPlanet}.

Measurement platforms have been collecting large-scale datasets of global
censorship to highlight its prevalence, as well as the mechanisms used to
implement it~\cite{filasto2012ooni, niaki2020iclab, Raman2020CensoredPlanet}.
These datasets consist of network probes conducted via remote vantage points;
the resulting measurements are typically compared against known `signatures',
based on anecdotes and heuristics, to confirm cases of
censorship~\cite{OONI-rules}. Unfortunately, censorship detection via such
heuristics can be quite brittle, because censors {\em adapt}; specifically, they
routinely change their blocking behaviors to evade fixed rules and heuristics,
rendering these rule-based approaches far less effective and
accurate~\cite{Tripletcensors, USESEC21:GFWatch}. We further investigate this
situation in~\sectionref{sec:motivation} by analyzing two large-scale censorship
datasets and find that they disagree considerably on the blocking status (i.e.,
`anomaly'~\footnote{The term ``anomaly'' refers to a network probe in which
censorship is suspected. According to these platforms themselves, this `anomaly'
may or may not be instances of actual censorship~\cite{OONI_qa}.}) of many
domain names despite employing similar heuristic approaches.

Given the inconsistencies we observe between heuristics and the need for
accurate censorship detection, we aim to explore more robust, automated
approaches. Machine learning (ML) methods are a potentially compelling and
sensible alternative in this context, due to the plethora of measurement data,
the existence of patterns that can be discovered, and in some cases even the
existence of labels. Furthermore, recent advances in ML have enabled efficient,
large-scale and interpretable anomaly detection, with promising results in
adjacent domains, such as
malware~\cite{arp2014drebin,ho2019detecting,zhu2016featuresmith} and
disinformation detection~\cite{hounsel2020identifying}. The availability of
large-scale censorship datasets, along with the existence of easy-to-use, yet
scalable ML methods compels us to investigate \textit{whether ML models can
assist in detecting DNS-based censorship}. Specifically, by applying different
ML models to datasets collected by DNS censorship measurement platforms, we
examine \textit{whether we can successfully identify instances of Internet
censorship while reducing dependence on hardcoded heuristics.}

The first step in training reliable ML models in any new domain is data
engineering; that is, obtaining clean data in the appropriate format. For this
study, we cleaned and formatted millions of probes/records~\footnote{We use
probes and records interchangeably throughout.} over 7 months from the two
prevailing large-scale measurement platforms, Satellite \cite{Satellite}
(maintained as part of the Censored Planet
platform~\cite{Raman2020CensoredPlanet}) and OONI~\cite{filasto2012ooni}. We
focus our analysis on two countries: China and the US (\sectionref{sec:method}).
Unfortunately, we discovered that the measurement data from Satellite and OONI
could not directly be used to train ML models, due to challenges such as
inconsistent data formatting, the presence of irrelevant features, and
incomplete records (e.g., failed measurements). Our \textbf{first contribution}
was thus to transform these datasets into a \emph{curated format amenable to
ML}, generating two datasets with over $3.5$M records from Satellite and $1.2$M
records from OONI~(\sectionref{sec:method}). These datasets will be open-sourced
for the community's use.

After performing appropriate data engineering, we study the feasibility of ML
models for DNS censorship detection. We consider models from the two main ML
paradigms, \emph{supervised} and \emph{unsupervised} learning. \emph{Supervised
models} require labeled data and we use labels obtained from the accompanying
heuristics for each data source to label each record. Additionally, when it
comes to measuring DNS censorship in China, the GFWatch~\cite{USESEC21:GFWatch}
platform presents an alternative source of labels for existing probes. This
third dataset enables us to make an important and new methodological innovation
when using ML for censorship detection: {\em training models on the same
records, with labels obtained from different sources} enables higher-confidence
detection, even in the absence of ground truth labels. In the context of
\emph{unsupervised models}, we extend this approach to more accurately identify
``clean'' (i.e., uncensored) records, which we then use to build a model of
``normal'' behavior.

Our \textbf{second contribution} is thus a thorough analysis of ML models'
ability to learn existing censorship detection rules, and find new instances of
(suspected) censorship in large-scale datasets
(\sectionref{subsec:results_original_rq}). The lack of ``ground-truth''
information about the censorship of a given probe is challenging to train ML
models that act as perfect indicators of censorship. Regardless, we show that ML
models enable the discovery of new instances and signatures of censorship at
scale, providing a valuable additional perspective for censorship detection. Our
key research questions and results are as follows:

\myparab{RQ 1.} \emph{\textbf{Can supervised models learn anomaly detection
heuristics employed by different measurement platforms?}} Yes. Given data with
anomaly labels assigned by a measurement platform's heuristics, optimal
supervised models can achieve true positive rates (TPR) in excess of $90\%$
while maintaining low false positive rates (FPR) below $5\%$
(Table~\ref{tab:supervised_all_results}). This demonstrates that there is a
generalizable signal of anomalous behavior in the records that is highly
correlated with the given labels that can be learned, regardless of whether the
labels themselves are accurate. This is a preliminary step towards using ML
to supplement heuristics, showing that automated ML models can perform on par
with manually derived heuristics.

\myparab{RQ 2.} \emph{\textbf{Can supervised models infer DNS censorship using
records and labels collected from independent sources?}} Yes. Using GFWatch
labels for records from both Satellite and OONI, we find that supervised models
can obtain high TPRs ($99.4\%$ for Satellite and $86.7\%$ for OONI) with respect
to these labels as well. This result highlights the richness of the censorship
signal contained within these probes and the methodological soundness of using
multiple labels for a given record. Since GFWatch labels only contain true
positives by construction, the positive instances identified by these models are
extremely likely to be true instances of censorship. Our \emph{post facto}
analysis shows these supervised models find instances of non-censorship and
localized censorship missed by GFWatch since the platform does not guarantee a
\emph{zero} FNR.

\myparab{RQ 3.} \emph{\textbf{Can unsupervised models discern censorship
instances that existing, state-of-the-art heuristics miss?}} Yes. Our
unsupervised models, trained on clean records that unlikely contain any
censorship instances, are able to detect anomalous records with a high level of
agreement with existing heuristics (Table~\ref{tab:unsupervised_all_results}).
Further, an in-depth case analysis of the disagreements between model
predictions and the anomaly status determined by the heuristics shows these
often occur in corner cases missed by heuristics.
For instance, among false negative cases, unsupervised models can
discern instances of temporarily accessible domains due to firewall failures,
which GFWatch is not designed to detect. Among false positives, we find
censorship instances occasionally missed by all heuristics due to
their reliance on hard-coded rules. These results demonstrate how unsupervised
models can complement existing methods to create more comprehensive censorship
detection systems.

Our analysis in~\sectionref{sec:results_in_depth} shows that predicted
censorship instances can be mapped back to their original records, revealing
\textit{hundreds of injected fake IPs} that can serve as blocking signatures to
confirm DNS censorship currently overlooked by heuristics. Feature importance
analysis shows that the most important features are interpretable, and can be
explained from first principles, increasing our confidence that the models
learned automatically are relying on the appropriate signals for detecting
censorship. In addition, we conduct a longitudinal comparison of DNS censorship
predicted by unsupervised models when trained and tested separately on OONI and
Satellite data. Our methods show a significantly greater agreement of censored
domains between OONI and Satellite compared to the original heuristic-based
labels.

These results collectively demonstrate ML models' ability in learning existing
anomaly detection heuristics at scale, even when the heuristics used to label
the data are independent of the particular measurement dataset. This observation
makes it possible to ensemble models for detecting censorship. We discuss some
future directions in~\sectionref{sec:discussion}. We
open-source~\footnote{\url{https://github.com/noise-lab/automated-dns-censorship}}
our code, datasets, and a dashboard to stimulate further ML-based censorship
detection research.

\section{Motivation}
\label{sec:motivation}

The Domain Name System~(DNS) plays a foundational role on the
Internet~\cite{rfc1035} by translating human-readable domain names to
corresponding IPs. DNS is thus necessary for the initiation of almost every
online communication. The insecure design of DNS~\cite{rfc1035} makes it a
popular means for censorship~\cite{Aryan:2013, Iris, USESEC21:GFWatch,
Padmanabhan2021AMV}. State-sponsored censors can use DNS to point users to
invalid IPs or redirect them to a blockpage by modifying DNS resource records at
ISP-provided resolvers~\cite{Iris, Padmanabhan2021AMV, hoang:2019:measuringI2P,
hoang2022measuring, USESEC21:GFWatch, Nourin2023:TMC}.

\subsection{Measuring DNS Censorship is Challenging}

\begin{figure*}[!htbp]
    \includegraphics[width=\textwidth]{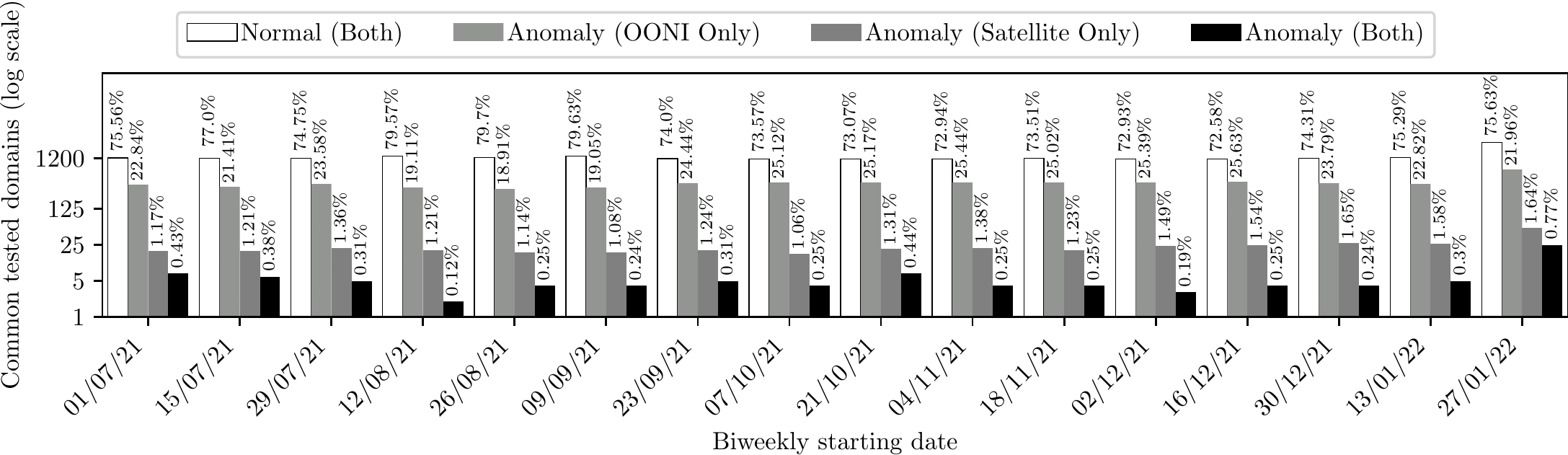}
    \caption{A longitudinal comparison of China's DNS censorship measured by
     OONI and Satellite.}
    \label{fig:ooni_cp_compare}
\end{figure*}

As a countermeasure, censored users may rely on public resolvers (e.g.,
\texttt{9.9.9.9}) to bypass manipulation occurring at default DNS resolvers,
such as the ones operated by their ISPs. Upon detecting censored domains from
users' queries, an on-path blocking system however can still inject forged
responses~\cite{Satellite, Iris, niaki2020iclab}. China, for example, is known
to use on-path DNS filtering systems~\cite{Sparks:CCR2012, holdonDNS,
USESEC21:GFWatch}.

Despite its prevalence, DNS manipulation is challenging to measure. A common
challenge faced by measurement platforms concerns drawing inferences about
whether censorship is taking place. For example, in the case of DNS, a
censorship event, such as via manipulation or redirection, can sometimes be
difficult to detect reliably because DNS responses vary depending on time and
geography as part of normal operations. As a result, even the most advanced and
reliable heuristic-based approaches in DNS censorship detection may still result
in false negatives, i.e., missing censorship instances. To this end, we explore
how these challenges can cause censorship measurement systems to reach
inconsistent conclusions about DNS manipulation
(\sectionref{subsec:heuristics}). Furthermore, heuristic-based approaches that
prioritize high precision require significant effort and resources, as they
necessitate ongoing censorship confirmation procedures such as the verification
of censorship instances with local advocacy groups in censored locations, which
can be both time-consuming and costly. These outstanding shortcomings lead us to
consider, in this paper, whether and how machine-learning models could improve
upon these existing rule-based systems (\sectionref{subsec:ml}).

\subsection{Rule-Based Heuristics Are Inconsistent}
\label{subsec:heuristics}

Different tools have developed static, rule-based heuristics to detect
censorship. Unfortunately, these heuristics (and the lack of ground truth
labels) have led to inconsistent inferences among different heuristic
techniques. For example, one approach to detect DNS censorship is to compare
responses obtained via a vantage point being tested and via a control server
located in a network where it is believed that censorship is not occurring. This
approach can be effective when forged responses contain static records that have
been known to be invalid (e.g., private non-routable IPs, NXDOMAIN) or point to
blockpages~\cite{rfc8020, syriacensorship, Aryan:2013, Padmanabhan2021AMV,
Nourin2023:TMC}. Unfortunately, this heuristic results in erroneous conclusions
when forged responses contain dynamic records created by censors to confuse
detection. China's Great Firewall (GFW) engages in such
behavior~\cite{Tripletcensors, USESEC21:GFWatch}. To show the impact of such an
adversarial behavior on DNS censorship detection, we performed a longitudinal
comparison between DNS censorship measurements collected by OONI and Satellite.
We analyzed measurements for China collected by both platforms over 32 weeks,
from July 1, 2021 to February 9, 2022~\footnote{Both platforms have been
collecting datasets prior to this period, yet, to ensure consistency, we opt to
analyze data from July 2021 because this is the first full month after the
Censored Planet team made the latest modifications to Satellite
v2.2~\cite{Satellitev2.2}.}.

Although both platforms test the same list of domains curated by the Citizen
Lab~\cite{CitizenLabtestlist}, they have different testing frequencies. We thus
perform our comparisons using biweekly time intervals to (1)~make the sets of
results comparable; and (2)~ensure that the set of domains tested overlaps as
much as possible. When the same domain is tested multiple times over two weeks,
its blocking status is determined by the blocking status of the majority of all
tests.

Figure~\ref{fig:ooni_cp_compare} shows around 1.6K common domains tested
biweekly by both platforms (note that the y-axis is in log-scale for better
visibility). Although both platforms agree largely on domains that are
\emph{not} censored ($75\%$ of tested domains), the number of `DNS anomaly'
cases (i.e., a potential sign of DNS manipulation) is relatively small ($<0.5\%$
most of the time). Such discrepancies exist because each platform has a
different design and employs different `anomaly' inference heuristics, and thus
being susceptible to dynamic and adversarial filtering behaviors of censors like
China's GFW.

To measure DNS censorship, both platforms issue DNS lookups for test domains
from their control servers and vantage points. DNS responses are then compared
to mark anomaly cases (i.e., likely censorship) if significant inconsistency
exists among these responses. However, the characteristics of vantage points
used by these platforms are inherently different since Satellite leverages open
DNS resolvers whereas OONI relies on local volunteers.

Some probes conducted from residential networks may experience localized
filtering policies (e.g., parental controls, school or corporate firewalls).
This can cause these platforms to arrive at different conclusions when inferring
country-level censorship. We sample one month (July 2021) of OONI data from
China, which performs DNS censorship~\cite{Tripletcensors, USESEC21:GFWatch}. We
cross-validate DNS anomalies observed by OONI with data from
GFWatch~\cite{USESEC21:GFWatch}, a measurement system that focuses primarily on
monitoring DNS filtering in China. We find that many of OONI's DNS anomalies are
due to probes conducted in autonomous systems (ASes) allocated with a small
number of IPs (see Table~\ref{tab:ooni_one_month},
Appendix~\ref{appsec:additional_results}).

\subsection{Machine Learning Models May Help}
\label{subsec:ml}

Machine learning has emerged as an effective approach to scale up the detection
and classification of vulnerabilities in many security~\cite{ho2019detecting,
zhu2016featuresmith, arp2014drebin, hounsel2020identifying,polverini2011using}
and networking~\cite{mirsky2018kitsune,yang2020feature,aceto2015internet}
contexts, including malware, network anomaly and disinformation detection. Both
supervised models~\cite{alpaydin2020introduction}, which use labeled data for
training, and unsupervised models~\cite{aggarwal2017introduction} for detecting
deviations from `normal' behavior have been used. Tree-based classifiers work
well and are interpretable~\cite{ho2019detecting,hounsel2020identifying}. Recent
work has also focused on creating measures of interpretability for features that
are model-agnostic and suited for applications beyond computer vision and
language~\cite{guo2018lemna,lundberg2017unified}. These developments, together
with the recent availability of datasets from measurement platforms such as OONI
and Censored Planet, motivate us to investigate whether ML can help automate
censorship detection using trained models.

\section{Method}
\label{sec:method}

We next describe how we curate training datasets to train ML models from the
Satellite and OONI, and the ML pipeline that uses supervised and unsupervised
models to detect DNS censorship.

\subsection{Datasets}
\label{sec:datasets}

To the best of our knowledge, active platforms that are still collecting and
publishing global censorship data at the time of writing include
OONI~\cite{filasto2012ooni} and Censored Planet~\cite{Raman2020CensoredPlanet},
whose datasets we use for our analysis in this study. Many other studies have
investigated regional censorship~\cite{Aryan:2013, pakistancensorship,
khattak.2014.isp, Yadav2018, Bock2020DetectingAE, Nourin2023:TMC}.

\myparab{Open Observatory of Network Interference
(OONI)~\cite{filasto2012ooni}.} Launched in 2012, OONI is one of the earliest
platforms developed to measure global censorship. The OONI Probe software is
built with different modules running on a volunteer's device. These modules test
the connectivity of websites, instant messaging services, and
censorship-circumvention applications. To date, OONI volunteers have conducted
986.6M measurements from 24.2K network locations in 200 countries. Numerous
regional censorship studies have been conducted based on this massive
dataset~\cite{xynou_how_2021, xynou_zambia_2021, xynou_uganda_2021,
basso_measuring_nodate, xynou_russia_2021, xynou_new_2022}.

\myparab{Censored Planet~\cite{Raman2020CensoredPlanet}.} Unlike platforms that
rely on volunteers~\cite{filasto2012ooni} or dedicated vantage
points~\cite{niaki2020iclab}, Censored Planet employs public infrastructure and
uses established measurement techniques (i.e., Satellite~\cite{Satellite},
Augur~\cite{Pearce2017AugurID}, Iris~\cite{Iris}, Quack~\cite{Quack}, and
Hyperquack~\cite{Raman2020MeasuringTD}). The platform curates several datasets
that collectively cover both application and network-layer censorship. We use
data from Satellite, which is based on Iris~\cite{Iris}. Satellite DNS
measurements are performed twice per week using public resolvers in more than
170 countries. Since its initial launch, Satellite has gone through several
revisions to improve accuracy and efficiency~\cite{Satellitev2.2}.

\myparab{GFWatch~\cite{USESEC21:GFWatch}.} China has perhaps the most
sophisticated censorship infrastructure of any nation-state, which motivated the
development of GFWatch for determining censored domains in China. The platform
performs daily large-scale DNS measurements for more than 400M domains using its
own machines located at both sides of the ``Great Firewall'' (GFW), continuously
monitoring the GFW's DNS filtering behavior. Unlike OONI and Satellite, the use
of control machines at different locations enables GFWatch to ensure that the
domains it determines to be censored, definitely are at a national-level.
Moreover, censored domains detected by GFWatch are frequently shared with owners
of blocked domains and local Chinese advocacy groups to cross-check the results
using their own independent testing infrastructure located in multiple network
locations across the country. {\em We thus use GFWatch as an additional source
of labels for records from both Satellite and OONI}, providing a label set with
no false positives with respect to ground truth national-level censorship. We,
however, note that GFWatch may still miss instances of local or regional
censorship, and is thus not free of false negatives (\sectionref{sec:results}).

\subsection{Data Labeling and Cleaning}
\label{subsec:data_curation}

Training first requires datasets to have minimal label errors and contain
anomalies of the type we are hoping to detect. Table \ref{tab:dataset_details}
summarizes these datasets.

\begin{table}
	\centering
	\resizebox*{\columnwidth}{!}{
	\begin{tabular}{l|llll}
	\toprule
				                  & Satellite (CN)    & Satellite (US)& OONI (CN) & OONI (US)  \\
	\midrule
	Initial                       & 1.85M             & 1.74M         &   937K    & 28M      \\
	\textit{sans} errors          & 1.76M             & 1.73M         &   671K    & 590K      \\ %(sampled)
	Clean                         & 1.14M             & 1.54M         &   496K    & 590K       \\
	Anomalous                     & 618K              & 189K          &   175K	  & 0K    \\
	\midrule
	Initial Features              & 68                & 68            &   17      & 17           \\
	Post conversion               & 1633              & 1633          &   77      & 77     \\
	\bottomrule
	\end{tabular}
	}
\caption{Number of records and features in curated datasets.}
\label{tab:dataset_details}
\end{table}

\myparab{Initial dataset and labels.} We examine data (\emph{records}) from
China and the United States (US) from July 1, 2021 to February 9, 2022. GeoIP
information of both OONI and Censored Planet data is inferred from the MaxMind
dataset. We use \emph{labels} from Satellite~\footnote{While the `anomaly' field
is now deprecated, it is available for the Satellite records within our time
frame of investigation and we use it for data cleaning as well as testing model
performance, with the caveat that their reliability is limited.}, OONI, and
GFWatch as an alternative label source in the case of China. We note that
records marked as ``anomalies'' \emph{may not} be censored~\cite{OONI_qa} since
ground truth about censorship is challenging to obtain. This motivates us to
train multiple models with different labels for the same records, to obtain
higher confidence predictions and to perform case analysis over the
disagreements.

We choose China as it has been shown to be one of the most sophisticated censors
on the Internet. In particular, the country's DNS blocking mechanisms have
continuously changed over the past two decades to hinder straightforward
detection and circumvention~\cite{USESEC21:GFWatch}. If our models work well on
China, they will also likely be capable of learning censors' signatures in other
countries and adapt to changes in censors' behavior over time. In addition,
GFWatch is available as a highly credible alternative label source, allowing us
to train our models with different labels for the same records, each reflecting
a partial view of the ground truth.

Note that, using GFWatch labels for Satellite records is not perfect due to
discrepancies in measurement frequency and probing times. There may be periods
in which a newly censored domain is probed by Satellite, but not yet probed by
GFWatch. A machine learning model acting on Satellite records may thus identify
the domain as censored even if it is not labeled as such, and vice versa. Since
the blocking status of domains change relatively infrequently, this misalignment
however has no significant impact on our study.

Since the US does not perform DNS censorship, the data collected by both OONI
and Satellite makes the US an ideal candidate as a control country. Some US
probes may experience DNS filtering due to corporate firewalls or parental
controls as opposed to nation-state censorship since the US government does not
force ISPs to block websites~\cite{FreedomHouse_US}. To ensure our US data is
``clean'', we removed such anomalies by using only probes whose control and
returned AS numbers of each probe are the same. Moreover, DNS filtering by
corporate firewalls and parental controls often has unique signatures (e.g.,
about 600 IPs of Cisco filters found in~\cite{USESEC21:GFWatch}) are also used
to remove anomalies due to residential and corporate filtering.

\myparab{Data sampling and balancing.} For countries with large Internet
infrastructures like China and the US, there are thousands of open resolvers
that Satellite can use. These resolvers are probed twice a week with thousands
of domains from the Citizen Lab test lists~\cite{CitizenLabtestlist}. We retain
probes collected from ASes with large Internet-population ratios based on the
RIPE Atlas Population Coverage data~\cite{RIPE-PopCover}. This yields a
representative dataset while maximizing coverage of the Chinese and US
population. For OONI, the amount of probes collected in the US is
disproportionately large compared to China. We thus opt to downsample the US
dataset uniformly at random to ensure a comparable ratio of US and CN probes to
those from Satellite. Since there are more than one million records in our final
training dataset, random selection has a negligible effect on the models'
performance. While there is imbalance within the datasets, it does not pose a
significant concern as censored instances constitute at least 10\% of the clean
instances for all datasets employed, amounting to millions of records.

\myparab{Verifying probe validity.} Neither OONI nor Satellite have control
over the network environment of their volunteers and remote vantage points.
Thus, tests can fail for reasons other than censorship, including
misconfiguration or failure.  For Satellite, we exclude probes if:
(1) the initial control query fails to resolve,
(2) the record shows \textit{zero} control resolvers with a valid response for
the test domain, or
(3) the total number of DNS responses received for the test domain is less than
	or equal to two across all countries during a single probing period (i.e.,
	the test domain was likely not active at that moment).
For OONI, we exclude probes if
(1) its control test failed,
(2) the ASN for the testing client or the client DNS resolver is missing or
invalid (i.e., AS0), or
(3)	the \texttt{body\_proportion} value between the control and response HTML
bodies is invalid, likely due to the inactivity of the tested website. Moreover,
we only retain OONI records that are marked either as `accessible' or
`DNS-tampered' since our main focus is on DNS censorship.

\myparab{Classifying probes as ``clean'' or ``anomalous''.} Unsupervised models
need ``clean'' records that are unlikely to be censored. Supervised models need
to be trained on a mixed dataset with both ``clean'' and ``anomalous'' records
(probes with unclear censorship status). For Satellite, we classify a probe as
``clean'' only if it satisfies all of the following four conditions. (1) It is
not marked as ``anomaly'' by Satellite's heuristics and not marked as censored
by GFWatch (China only). (2) It does not contain features that could indicate
censorship (i.e., a failed terminal control query). (3) Every control query does
not contain a connection error. And, (4) the ground-truth ASN of the test domain
appears in the Censys/MaxMind lookup of the IPs in the probe's \texttt{response}
section.

For OONI, a record is considered as ``clean'' if (1) it is not marked as ``DNS
tampered'' by OONI and is not marked as censored by GFWatch (China only), and
(2) the returned IP(s) in its DNS response is consistent with the one(s)
observed at OONI's control server. Records eliminated in this process are
classified as ``anomalous'' and not used when training unsupervised models. Via
this process, it is essential to stress our design choice of using two different
label sources for each dataset to more accurately identify ``clean'' records.

\subsection{Machine Learning Pipeline}
\label{subsec:ml_pipeline}

\begin{figure}[t]
    \includegraphics[width=\columnwidth]{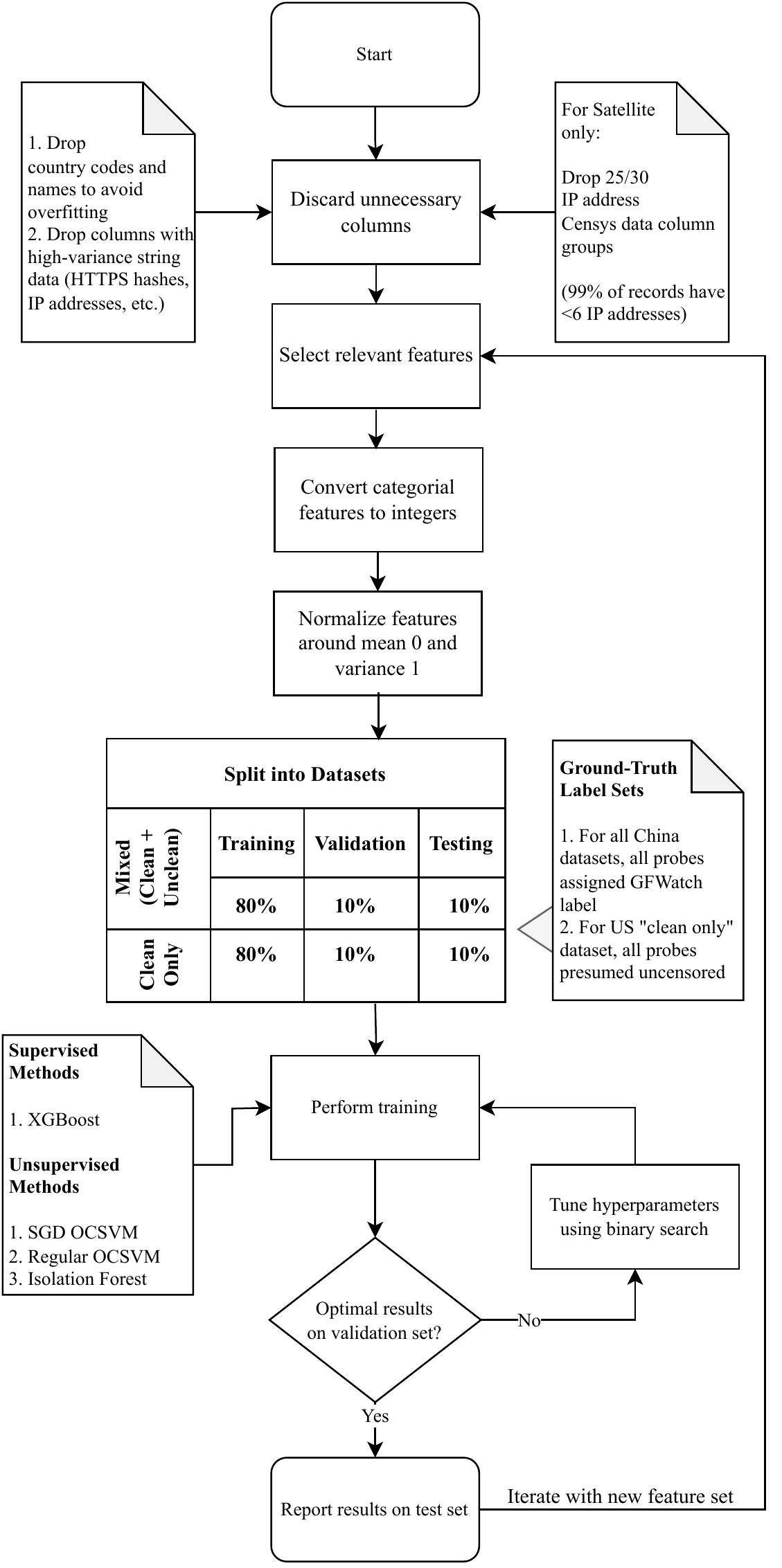}
    \caption{The machine learning pipeline for anomaly detection using Satellite/OONI datasets.}
    \label{fig:ml_pipeline_diagram}
\end{figure}

\myparab{Feature selection.} Both datasets include heterogeneous features such
as IPs and domains. Since only a few records share the same values, we remove
these features to reduce the chance of overfitting. For the remaining features,
we keep the numerical ones (e.g., measurement time length) and convert the rest
into categorical variables (e.g., response codes). We group features that are
strings, such as \texttt{http\_experimental\_failure} into error groups before
applying one-hot encoding. The features are then scaled and normalized.

In addition, we also verify whether there is any performance gain if features
(1) deemed as less-relevant to national-level DNS censorship by domain experts
and (2) highly correlated with one another, are excluded. We compare the
performance of the models with and without dropping those features and find
negligible performance difference. Tables~\ref{tab:ooni_features}
and~\ref{tab:satellite_features} in Appendix~\ref{appsec:data_description} list
features extracted from both datasets and their descriptions. Since we are also
committed to human-interpretability of our models, we opt to not use complex
feature transformations.

\myparab{Dataset curation.} We split the records into training, validation, and
testing datasets as shown in Figure~\ref{fig:ml_pipeline_diagram}. For
\emph{supervised models}, we create mixed datasets that contain \emph{both}
``clean'' and ``anomalous'' data (Table \ref{tab:supervised_all_results}). For
\emph{unsupervised models}, the training dataset only contains ``clean'' data,
while the validation and test datasets use all the probes (Table
\ref{tab:unsupervised_all_results}). The two sets of records for China (OONI and
Satellite) are labeled twice, once with their own labels and once using GFWatch,
leading to two separate models. The validation data is used for hyperparameter
selection while the test data is held-out and for performance evaluation. We
conduct three-fold cross-validation and observe only small changes to the
results.

\myparab{Model selection.} Due to the large scale of these datasets, we select
models that are both memory and computation efficient. For supervised models, we
used Gradient Boosted Trees (XGBoost)~\cite{friedman2001greedy}, an ensemble
learning algorithm based on distributed gradient-boosted decision trees. XGBoost
performed better than all alternatives we tested such as SVMs and Random
Forests. For unsupervised models, we chose One-Class SVMs
(OCSVMs)~\cite{scholkopf2002learning} and Isolation Forests
(IF)~\cite{liu2008isolation} because of their ability to train on large data
sets. OCSVM can run in linear time when using the stochastic gradient descent
(SGD) heuristic. We performed hyperparameter search over maximum tree depth and
tree number for XGBoost, the maximum number of iterations for OCSVM, and maximum
features, number of estimators, and contamination rate for IF. Our open-source
code has further details. While we considered implementing more complex models
such as neural networks, their higher training costs and unclear performance
improvement for tabular data over classical models~\cite{grinsztajn2022tree}
make them less than ideal. Our methods already achieve high TPR/TNR for the
purposes of our deployment.
	
\myparab{Evaluation metrics.} We use multiple metrics to evaluate each model due
to class imbalance across the dataset. In our terminology a ``positive'' label
implies the presence of an anomaly, possibly indicating censorship, while a
``negative'' label implies no censorship. With this in mind, we report TPR
(sensitivity), TNR (specificity), FPR, and FNR. Other metrics, such as the
area-under-curve (AUC) and precision/recall, are also used to evaluate models'
performance. We report overall accuracy but do not use it in our evaluation
schemes because of the class imbalance.

\section{Results}
\label{sec:results}

\begin{table*}
	\centering
	\resizebox{.75\textwidth}{!}{
		\begin{tabular}{llllllll} 
			\toprule
			Train/Val./Test Records & Train/Val./Test Labels & TPR& FPR& TNR & FNR& Acc.&Prec.\\
			\midrule
			Satellite (China, Mixed) & GFWatch  & \textbf{99.4} & 0.9 &99.1&0.6&  99.1         & 95.4 \\
			Satellite (China, Mixed) & Satellite  & \textbf{100} & 0.00 &100&0.00&  100         & 100 \\
			\midrule
			OONI (China, Mixed)      & GFWatch & \textbf{86.7}    & 0.6&99.4&13.3& 96.1  &   98.1          \\
			OONI (China, Mixed)      & OONI              & \textbf{99.8} & 0.2  &99.8&0.2& 99.8 & 99.1   \\
			\bottomrule
	\end{tabular}}

	\caption{Performance of XGBoost, which is the most optimal \emph{supervised}
	model for both Satellite and OONI.}
	\label{tab:supervised_all_results}
\end{table*}

\begin{table*}
	\centering
	\resizebox{.9\textwidth}{!}{
		\begin{tabular}{lllllllll}
			\toprule
			Train Records & Val./Test Records & Val./Test Labels &TPR& FPR& TNR & FNR& Acc.&Prec.\\
			\midrule
			Satellite (China, Clean) & Satellite (China, Mixed)  & GFWatch & \textbf{99.1}&17.4&82.6& 0.9  & 85.3&52.7 \\
			Satellite (China, Clean) & Satellite (China, Mixed)  & Satellite & \textbf{100}&28.8&71.2& 0.00  & 74.3&29.6 \\
			Satellite (US, Clean) & Satellite (US, Clean)  & Satellite& \textbf{0.00} & 0.00 & 100 & 0.00  & 100 & 0.00\\
			\midrule
			OONI (China, Clean) & OONI (China, Mixed)  & GFWatch & \textbf{73.7}&10.0&90.0& 26.3  & 85.7& 72.4\\
			OONI (China, Clean) & OONI (China, Mixed)  & OONI & \textbf{88.8}&10.5&89.5& 11.1  & 89.3&68.9 \\
			OONI (US, Clean) & OONI (US, Clean)  & OONI & \textbf{0.00} & 0.00&100& 0.00  & 100&0.00\\
			\bottomrule
	\end{tabular}}
	
	\caption{Performance of OCSVMs and IFs, which are the most optimal
		\emph{unsupervised models} for Satellite and OONI, respectively.}
	\label{tab:unsupervised_all_results}
\end{table*}

We next evaluate the performance of both supervised and unsupervised ML models.
We first validate the use of ML for DNS censorship detection with respect to the
research questions laid out in~\sectionref{sec:intro}. We analyze the false
negatives and positives found by our models with respect to the 3 labeling
heuristics used, uncovering errors in the heuristics and discovering new
censorship instances. We then explore our results, with a focus on the
interpretability of the results via feature importance analysis, along with case
studies on new censorship instances found by our models.

\subsection{Evaluation at scale}
\label{subsec:results_original_rq}

\myparab{RQ1: Can ML models learn ``anomaly'' detection heuristics employed by
various measurement platforms?} We verify whether existing `anomaly' labeling
heuristics can be learned automatically without the rules being explicitly
specified. We find that among all \emph{supervised models}, XGBoost performs the
best (Table~\ref{tab:supervised_all_results}), achieving nearly perfect TPRs
($100\%$ for Satellite and $99.8\%$ for OONI) at very low FPRs ($0.0\%$ and
$0.2\%$ respectively). This demonstrates that heuristics can be learned without
rule specification. 

However, it is more interesting to verify if \emph{unsupervised models} can
learn heuristics without the use of labels during training, as this can enable
post-heuristic censorship detection systems. Unsupervised models
(Table~\ref{tab:unsupervised_all_results}) achieve high TPRs at the expense of
high FPRs, indicating that the calibration of our models tends to make them more
aggressive while determining potential instances of censorship. Regardless,
their ability to pick up anomalies as determined by heuristics without explicit
labels provided during training shows their effectiveness at creating a model of
``normal'' behavior.

We perform a sanity check of our unsupervised models by assessing their
performance against US records which contain \textit{zero} DNS censorship. This
is to ensure that our models are not badly calibrated and identify anomalies
where there are none. For both OONI and Satellite datasets, our models pass the
sanity check against the clean US records with TNRs at $100\%$
(Table~\ref{tab:unsupervised_all_results}).

\myparab{RQ2: Can supervised models infer DNS censorship using records
and labels collected from independent sources?}
Censorship labels obtained from Satellite and OONI can be unreliable and
inconsistent (\sectionref{subsec:heuristics}). Leveraging the availability of
GFWatch for labeling probes with respect to national-level DNS censorship in
China \emph{with a FPR of zero}, we aim to create a more reliable labeled
dataset. This allows us to train supervised models at scale that have
trustworthy positive predictions, and also provides a unified way to compare DNS
probes of different varieties. Table~\ref{tab:supervised_all_results} shows the
performance of our supervised ML models on this curated dataset.

For both OONI and Satellite, the XGBoost models achieve relatively good
performance. However, a discrepancy exists between the datasets: for Satellite,
the best model achieves a very high TPR and low FPR, implying a high degree of
correlation between the information contained in the Satellite records and
GFWatch labels. On the other hand, the model using OONI probes has a lower TPR
(86.7\%), implying the existence of a significant fraction of probes where
GFWatch indicates the domain should be censored but our model marks probes from
those domains as uncensored. Overall, however, this shows we can train models
to extract censorship signals correlated with different label sets, allowing for
the study of discrepancies and higher confidence predictions with ensembling.

\emph{\textbf{False Negative Analysis of OONI probes with GFWatch labels:}} We
perform a thorough manual investigation of the probes belonging to the 13.3\% of
cases marked as false negative, i.e., probes that our supervised model trained
on OONI using GFWatch labels predicted as \emph{not} censored but was marked as
censored by GFWatch labels. Intriguingly, we find that more than 67\% of these
false negatives are \emph{truly negative} cases, i.e., \emph{not} censored
probes. Prior work has discovered that the GFW is sometimes overloaded and may
fail to inject forged DNS responses~\cite{USESEC21:GFWatch}. By manually
examining these FN cases, we verify that these probes could actually obtain the
correct IP(s) associated with their test domains.

For instance, this probe~\cite{OONI-supervised-corner-case} of
\texttt{www.washingtonpost.com} conducted in AS4134 (CHINA UNICOM) is marked as
`DNS tampering' by OONI and is also labeled as censored by GFWatch. However, the
resolved IP correctly points to AS20940 Akamai (the primary hosting provider for
Washington Post), indicating that the GFW failed to inject a forged response in
this case. Our model, thus, could correctly classify the DNS resolution of this
probe as \emph{not} censored.

In addition, localized censorship also contributes to the false negative cases.
They can be caused either by local filtering (e.g., corporate firewalls) or
server-side blocking~\cite{McDonald:2018:403}. We found several false negatives
that did not actually experience the Great Firewall's DNS censorship but were
redirected to blockpages due to web-application firewalls (e.g.,
\texttt{192.124.249.111}).

Overall, our ML models allow for finer-grained discovery of aspects of DNS
censorship. Even heuristics like GFWatch which deploy physical machines in
censorship regimes can have erroneous labels for specific probes due to their
reliance on aggregate trends.

\myparab{RQ3: Are unsupervised models capable of discerning censorship instances
and variations missed by existing heuristics?}
Table~\ref{tab:unsupervised_all_results} reports the performance of the most
optimal unsupervised models trained on both datasets' China records, labeled
with GFWatch's labels. With a TPR of $99.1\%$ and a FPR of $17.4\%$, a Linear
OCSVM model trained on Satellite provides relatively better performance in
comparison to the Isolation Forest (IF) model trained on OONI ($73.3\%$ TPR and
$10.0\%$ FPR). The latter, however, has a slightly higher accuracy ($85.7\%$ vs.
$85.3\%$). The positive instances identified by these models are validated using
GFWatch labels.

\emph{\textbf{False Negative Analysis:}} By examining the FN cases, we found
that more than half of them are \emph{truly negative}. In other words, the
``normal'' behavior that the unsupervised models have learnt could also
correctly discern corner cases due to temporary failures of the GFW or incorrect
geolocation of IPs. Thus, even though the domain is actually blocked in China,
individual probes may go through due to the GFW's temporal
failures~\cite{USESEC21:GFWatch} or incorrect location of testing vantage points
as inferred using MaxMind GeoIP service~\cite{Raman2020CensoredPlanet}.

\emph{\textbf{False Positive Analysis:}} The unsupervised models have a higher
number of false positives as verified by a domain expert, indicating
conservative boundaries for ``normal'' behavior. However, this aspect does lead
to a number of potential false positives with respect to the heuristics being
\emph{instances of possible censorship caught only by our model}, with all the
label sets missing them. For example, we find that some Satellite probes for
\texttt{messenger.com} occasionally undergo DNS censorship since the returned IP
points to a known fake IP used by the GFW. However, this website in general only
undergoes HTTP/HTTPS censorship, with rare instances when DNS filtering is also
activated. Our model is thus sensitive enough to uncover these rare events. In
addition, more than 10\% of the FP cases are due to the inactivity of the test
domains at probe time leading both Satellite and OONI to be unable to fetch
their websites.

The good performance and ability of unsupervised models at discerning DNS
censorship cases missed by existing labeling heuristics is encouraging for
future automated censorship detection efforts. Models with minimal labels are
able to find anomalies, which can aid censorship detection in regimes where it
is poorly understood.

\begin{figure}[t]
	\centering
	\includegraphics[width=\columnwidth]{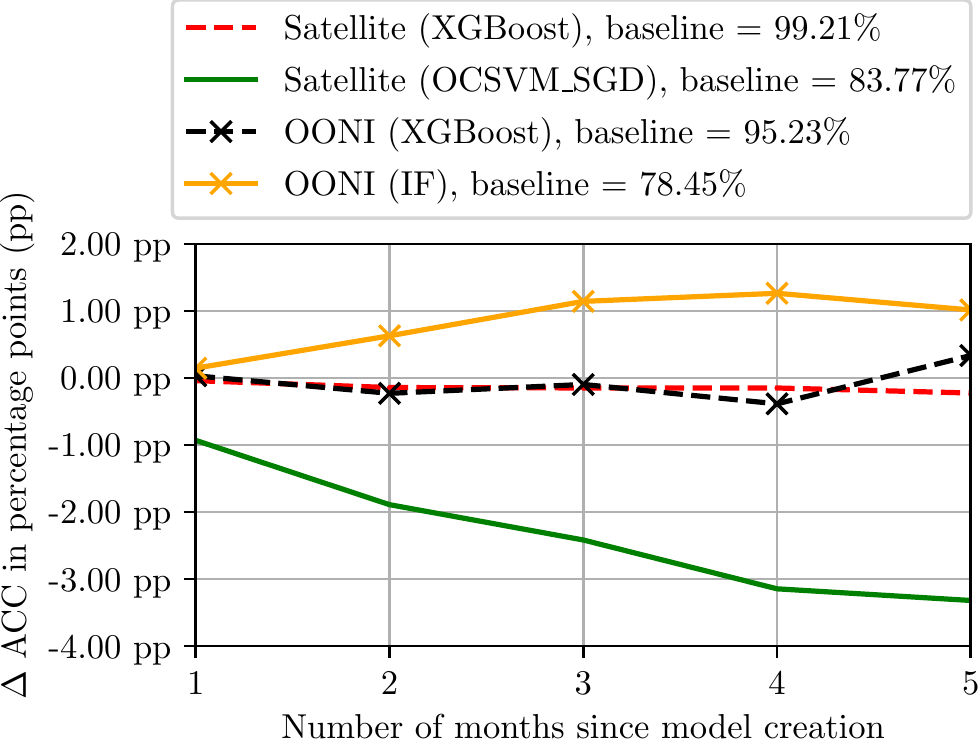}
	\caption{An analysis of ML models' performance change over time. The baseline performance number is the average accuracy of the most recent possible model (1 month).}
	\label{fig:temporal_supervised}
\end{figure}

\subsection{In-Depth Analysis of Results}
\label{sec:results_in_depth}
We now analyze the performance of our models further by interpreting the
features used, their performance variation over time, as well as the discovery
of new aspects of censorship behavior.

\noindent \textbf{Feature importance analysis.} To show how our supervised
models pick up signals used by heuristics, we apply the SHAP
method~\cite{lundberg2017unified} on the XGBoost for OONI
(Table~\ref{tab:ooni_fi}, Appendix~\ref{appsec:additional_results}).
Encouragingly, the model's most important features ranked by SHAP align well
with those used by the OONI's heuristics, including ASN and organization name of
the returned IP(s) in the DNS response. In particular, OONI compares the
returned IPs between the test and control queries, and concludes ``anomaly'' if
they do not match. Since an IP can be mapped to its respective AS information,
the supervised model is essentially learning this logic.

Similarly, our unsupervised models also make predictions based on a set of
important features that domain experts would typically utilize to confirm
censorship, e.g., ASN and organization name of returned IP(s) together with
other HTTP-related elements, including failure codes, and HTML body length to
confirm the blocking status. They also exhibit remarkable consistency with their
supervised counterparts in terms of the most important features.
Table~\ref{tab:satellite_fi} in Appendix~\ref{appsec:additional_results} shows
feature importance analysis for Satellite. Both models show promising learning
capability by properly identifying AS information as one of the most important
features. This is because AS-related features are more robust than IP
information alone since a website may be mapped to different IP, depending on
when and from where its DNS query is resolved~\cite{Hoang2020TheWI,
Hoang2021:IP-WF, Hoang2020:ASIACCS}.

\begin{figure*}[t]
	\includegraphics[width=\textwidth]{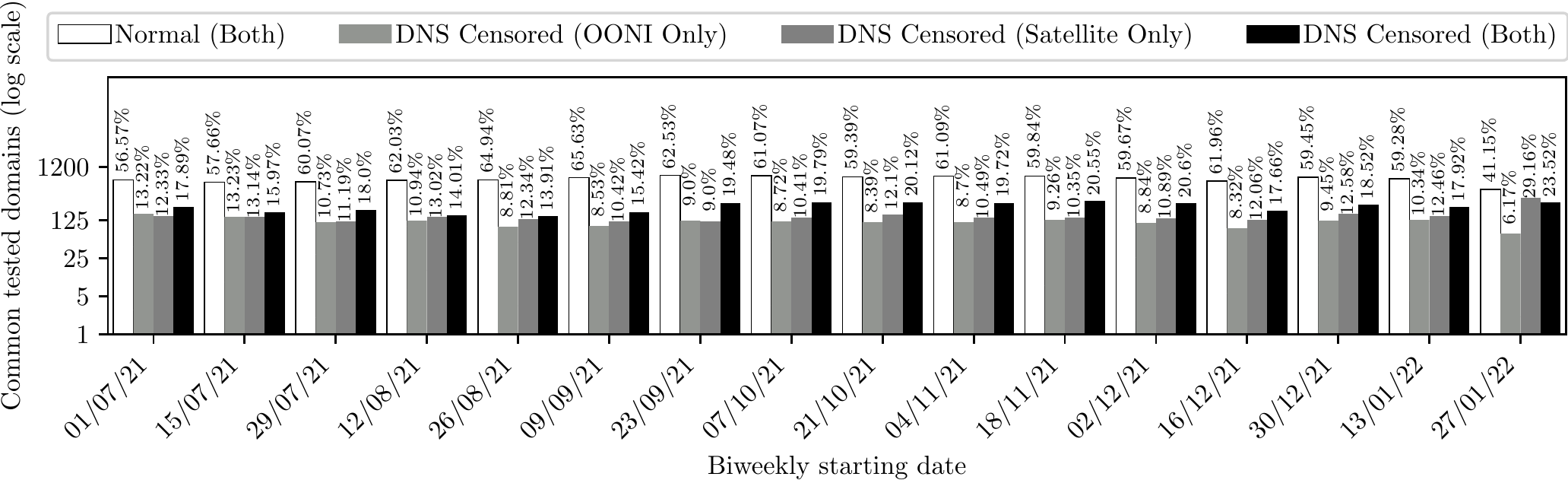}
	\caption{A longitudinal comparison of China's DNS censorship predicted using
	unsupervised models trained on OONI and Satellite data. We find much greater
	agreement among the predictions from the two sources of records when
	compared to using the original labels for the records.}
	\label{fig:prediction_ooni_vs_satellite}
\end{figure*}

\myparab{Model validity over time.} For our baseline experiments, we simply
combined data from the entire $7$-month time period we considered before
splitting it into training, validation and test data. The implicit assumption we
used here is that the data is i.i.d. over time. However, in real-world settings,
models can only be trained on currently available data to predict censorship for
future probes. Since censor behavior can change over time, the i.i.d. assumption
may be violated and models may need to be retrained frequently. In light of
this, we conduct experiments where we train models using a month of data, and
test them on all subsequent months. We then average the performance of all
models of a given age with respect to the test data and plot the results in
Figure~\ref{fig:temporal_supervised}. For Satellite models, both supervised and
unsupervised, we find that the performance decreases steadily over time as
expected, although the change is quite small. For OONI data, however, we observe
considerable variation over time, with performance even increasing for older
models. We believe this arises due to the scattershot nature of OONI data
collection from volunteers, as opposed to the regular probing done by Satellite.
This can lead to older models being more relevant even though more time has
elapsed. To summarize, our analysis indicates that, for both datasets, there
exists at least one trained model that exhibits robust performance as time
elapses. This serves as strong evidence that the models have successfully
learned transferable features, as opposed to features that are specific to the
context of the original datasets, and can generalize to future datasets
collected in different time-frames.

\myparab{DNS blocking signature discovery.} To confirm DNS censorship, platforms
such as OONI maintain lists of forged IPs~\cite{OONIIPs, OONI-rules}. However,
these lists are often static and small, leading to inadequate coverage and
potentially missing censorship cases. We thus perform a frequency analysis of
the associated IPs of probes~\footnote{Note that we excluded DNS response IP
addresses from the feature set to prevent overfitting, but we can still map the
predicted censorship instances back to the original records to examine the
associated DNS response IPs.} marked as censored by our models to find new fake
IPs that serve as DNS blocking signatures.

Table~\ref{tab:ooni_signatures} demonstrates that our models successfully
discovered hundreds of distinct, injected fake IPs that went unnoticed by OONI.
We also compare our results with those of GFWatch, which maintains a highly
comprehensive list of blocking signatures. Our model was able to identify
several new, previously undetected, injected ISP-specific censorship signatures
that are not picked up even by GFWatch. We provide a list of example signatures
found across models in Table~\ref{tab:ips}
(Appendix~\ref{appsec:additional_results}). Based on the results, we are
confident that these newly discovered DNS blocking signatures from our machine
learning models can significantly enhance the measurement platforms' ability to
accurately confirm cases of DNS censorship, providing more comprehensive
coverage. This, in turn, enhances the usefulness of the collected data for wider
public consumption.

\begin{table}
\resizebox{\columnwidth}{!}{
\begin{tabular}{llll}
\hline
\textbf{Model} & \textbf{\makecell{Train/Val./Test\\Labels}} & \textbf{\makecell{New Signatures\\ (missed by OONI)}} & \textbf{\makecell{New Signatures\\ (missed by GFWatch)}} \\ \hline
Supervised & GFWatch & 748 & 3 \\
& OONI & 747 & 2 \\ \hline
Unsupervised & GFWatch & 729 & 2 \\
& OONI & 728 & 1 \\ \hline
\end{tabular}}
\caption{New DNS blocking signatures uncovered by models when using OONI data.}
\label{tab:ooni_signatures}
\end{table}

\section{Discussion and Future Work}
\label{sec:discussion}

\myparab{Synthesizing disparate information sources.} An essential insight
obtained through this work is the effectiveness of combining disparate
information sources for censorship detection. Comparing the outputs of
heuristics and ML models led to the identification of new censorship instances.
Further, using records from one source and labels from another allows us to make
use of sources of records like Satellite where the labeling method has been
deprecated. Using unsupervised models, we are also able to consistently find
common domains for which both records from OONI and Satellite indicate possible
DNS censorship (Figure~\ref{fig:prediction_ooni_vs_satellite}). This allows us
to effectively bypass the lack of reliable labels from the data sources. It also
makes effective use of multiple sets of measurements, with the set of
overlapping domains being extremely likely targets of censorship. Aggregating
predictions over time also increases our confidence in whether a domain is
actually censored. Thus, synthesizing information from different models
(including heuristics), at different times, allows for reliable determination of
censorship trends.

\myparab{Generalization across countries.} An important direction for future
work is to validate if models trained on data from one country will generalize
to another. To test if our models will be effective at this, we dropped all the
country- and region-specific features such as ASN from our data before
retraining. As shown in Table~\ref{tab:satelliteasndropping}
(Appendix~\ref{appsec:additional_results}), even without these features, both
supervised and unsupervised models perform well. This is important in particular
for unsupervised models, since the lack of a system like GFWatch can then be
overcome by using a model of ``normal'' behavior obtained from other countries
to detect censorship in countries whose DNS filtering is poorly understood. On
another note, we choose China as our case study due to its extremely
sophisticated censorship behaviors, with constant churn in fake
IPs~\cite{GFWatch}. The effectiveness of our models there provides confidence
that they will work for other complex censorship regimes. Several others have
extremely simplistic blocking mechanisms that can be detected trivially. For
instance, Iran injects static private IPs from the range 10.0.0.0/8 while
Turkmenistan uses only one private IP (127.0.0.1) for DNS poisoning. We hope
that our paper inspires future work on using ML techniques to detect censorship
in other similarly complex censorship regimes such as Russia, but that is
unfortunately an engineering and research effort that is beyond the scope of
this single paper. 

\myparab{Limitations.} While acknowledging that our curated datasets inherit
some biases from OONI and Satellite labels, we adopts a three-pronged approach
to reduce dependency on their potentially noisy labels: by seeking label source
agreement, using reliable GFWatch labels, and employing unsupervised models with
high true positive rates. The prospect of advanced feature engineering holds
potential for incremental improvement. We anticipate further exploration of more
intricate data transformations and model architectures, such as ensembles, in
future research. Our rationale for employing `simple' features in this study is
to facilitate human interpretability of our models utilizing tools such as SHAP.
Using complex features would impede our capacity to compare our methodologies to
existing heuristics and thus provide a persuasive explanation for their ability
to provide superior performance results.

\myparab{Looking ahead.} We believe that future efforts to build censorship
detection systems should employ both heuristics and the outputs of trained ML
models. This is because hand-labeled methods like GFWatch are imperfect and
cannot cover all variations of censorship within each country. Moreover, since
`ground-truth' datasets like GFWatch do not exist for countries other than
China, our results indicate that unsupervised models can be used as a
general-purpose tool for discovering DNS censorship in countries around the
world. However, our analysis of the cases where inconsistencies arise between
heuristics and ML models reveals that the former act as an important backstop,
with our models sometimes missing `obvious' instances of censorship or being
overly aggressive in marking censorship. Overall, our results point to the need
for an ensemble of models and heuristics whose design is carefully calibrated by
expert supervision for reliable detection of censorship.

\section*{Acknowledgements}

We would like to thank all the anonymous reviewers for their thorough feedback.
We also thank Manan Goenka, and others who preferred to remain anonymous for
helpful suggestions to improve this paper. This work was supported in part by
DARPA through award HR00112190126.

\bibliographystyle{ACM-Reference-Format}
\bibliography{conferences, reference}
\appendix

\section{Feature Descriptions}
\label{appsec:data_description}
Tables~\ref{tab:ooni_features} and~\ref{tab:satellite_features} contain complete
lists of the features we extract from OONI and Satellite to train ML models,
with (*) indicating the features that we have tried excluding from our models.
Detailed descriptions of these features can be found in the OONI and Satellite
documentations~\cite{ooni_fd, satellite_fd}.

\begin{table*}[!htbp]
    \resizebox{\textwidth}{!}{
    \begin{tabular}{l|l|l|l|l}
    \toprule
    \textbf{Feature Name} & \textbf{\makecell{Feature\\Type}}  & \textbf{\makecell{One-Hot\\Encoding}} & \textbf{Derived} & \textbf{Description} \\
    \midrule

        Measurement\_start\_time (*) & continuous & N & N & Start time of the measurement \\
        probe\_asn               & discrete   & Y & N & ASN of the testing client \\
        probe\_network\_name     & discrete   & Y & N & Name of the network in which the testing client located \\
        resolver\_asn            & discrete   & Y & N & ASN of the client's DNS resolver\\
        resolver\_network\_name  & discrete   & Y & N & Name of the network in which the testing client located \\
        test\_runtime            & continuous & N & N & Runtime of the test \\
        test\_start\_time (*)       & continuous & N & N & Start time of the test \\
        dns\_experiment\_failure & discrete   & N & N & Whether the DNS experiment failed \\
        dns\_consistency         & discrete   & N & Y & Whether the DNS response is consistent with the one collected via the control server \\
        http\_experiment\_failure (*)& discrete   & Y & N & DNS is consistent but failed to initiate HTTP connection \\
        body\_length\_match      & discrete   & N & Y & Whether the length of HTTP body observed at the client side matches that at the control server \\
        body\_proportion         & continuous & N & Y & Proportion of the control and the response body \\
        status\_code\_match      & discrete   & N & Y & Whether the status code of HTTP response at client side matches with that at the control server \\
        headers\_match           & discrete   & N & Y & Whether the HTTP headers observed at the client side match with those at the control server\\
        title\_match             & discrete   & N & Y & Whether the page's title observed at the client side matches with that at the control server\\
        test\_keys\_asn          & discrete   & Y & N & ASN for the specific network experiment we are doing \\
        test\_keys\_as\_org\_name& discrete   & Y & N & Name of the network for the specific network experiment \\
    \bottomrule
    \end{tabular}}
     \caption{Description for the features extracted from OONI that are used in training our machine learning models.}
    \label{tab:ooni_features}
    \end{table*}

\begin{table*}[!htbp]
    \resizebox{\textwidth}{!}{
    \begin{tabular}{l|l|l|l|l|l}
    \toprule
    \textbf{Feature Name} & \textbf{\makecell{Feature\\Type}} & \textbf{\makecell{One-Hot\\Encoding}}   & \textbf{Derived} & \textbf{Repetitions} & \textbf{Description} \\
    \midrule
    untagged\_controls                     & Discrete              & N                                  & N                & NA                   & \makecell{True if all the type-A records returned by the control probe cannot be tagged by Censys/MaxMind} \\
    untagged\_response                     & Discrete              & N                                  & N                & NA                   & \makecell{True if all the type-A records returned by the response probe cannot be tagged by Censys/MaxMind} \\
    passed\_liveness                       & Discrete              & N                                  & N                & NA                   & \makecell{True if at least one control query on the test resolver was successful} \\
    connect\_error                         & Discrete              & N                                  & N                & NA                   & \makecell{True if connection errors occurred for all of the test queries} \\
    in\_control\_group                     & Discrete              & N                                  & N                & NA                   & \makecell{True if at least one control resolver had a valid response for the test   domain} \\
    excluded\_below\_threshold             & Discrete              & N                                  & N                & NA                   & \makecell{True if all of the answer IP addresses appear only for 2 domains or less} \\
    delta\_time                            & Continuous            & N                                  & Y                & NA                   & \makecell{Difference between start and end time of the probe} \\
    control\_response\_start\_success      & Discrete              & N                                  & Y                & NA                   & \makecell{True if the first control query returned a response} \\
    control\_response\_end\_success        & Discrete              & N                                  & Y                & NA                   & \makecell{True if the last control query returned a response} \\
    control\_response\_start\_has\_type\_a & Discrete              & N                                  & N                & NA                   & \makecell{True if the starting control test returns at least one type-A record} \\
    control\_response\_start\_rcode        & Discrete              & Y                                  & N                & NA                   & \makecell{Starting control query response code mapped to RFC 2929} \\
    control\_response\_end\_has\_type\_a   & Discrete              & N                                  & N                & NA                   & \makecell{True if the ending control test returns at least one type-A record} \\
    control\_response\_end\_rcode          & Discrete              & Y                                  & N                & NA                   & \makecell{Ending control query response code mapped to RFC 2929} \\
    test\_query\_successful                & Discrete              & N                                  & Y                & NA                   & \makecell{True if at least one of the four queries with the test domain successfully   returns a response} \\
    test\_query\_unsuccessful\_attempts    & Continuous            & N                                  & Y                & NA                   & \makecell{Indicates the number of failed attempts to ask the DNS server about the   test query (min 0, max 4)} \\
    test\_noresponse\_1\_has\_type\_a      & Discrete              & Y                                  & N                & NA                   & \makecell{True if the first response attempt returns at least one type-A record; -2   if unsuccessful} \\
    test\_noresponse\_1\_rcode             & Discrete              & Y                                  & N                & NA                   & \makecell{The first response's return code mapped to RFC 2929; -2 if unsuccessful} \\
    test\_noresponse\_2\_has\_type\_a      & Discrete              & Y                                  & N                & NA                   & \makecell{True if the second response attempt returns at least one type-A record;   -1 if unnecessary; -2 if unsuccessful} \\
    test\_noresponse\_2\_rcode             & Discrete              & Y                                  & N                & NA                   & \makecell{The second response's return code mapped to RFC 2929; -2 if unsuccessful;   -3 if unnecessary} \\
    test\_noresponse\_3\_has\_type\_a      & Discrete              & Y                                  & N                & NA                   & \makecell{True if the third response attempt returns at least one type-A record; -1   if unnecessary; -2 if unsuccessful} \\
    test\_noresponse\_3\_rcode             & Discrete              & Y                                  & N                & NA                   & \makecell{The third response's return code mapped to RFC 2929; -2 if unsuccessful;   -3 if unnecessary} \\
    test\_noresponse\_4\_has\_type\_a      & Discrete              & Y                                  & N                & NA                   & \makecell{True if the fourth response attempt returns at least one type-A record;   -1 if unnecessary; -2 if unsuccessful} \\
    test\_noresponse\_4\_rcode             & Discrete              & Y                                  & N                & NA                   & \makecell{The fourth response's return code mapped to RFC 2929; -2 if unsuccessful;   -3 if unnecessary} \\
    test\_response\_has\_type\_a           & Discrete              & Y                                  & N                & NA                   & \makecell{If at least one test response returned without error, mark as 1 if at   least one type-A record is returned or 0\\if no records are returned; if no   test responses were successful, mark as -1} \\
    test\_response\_rcode                  & Discrete              & Y                                  & N                & NA                   & \makecell{The successful response's return code mapped to RFC 2929, or -2 if there   were no successful responses} \\
    test\_response\_IP\_count              & Continuous            & N                                  & Y                & NA                   & \makecell{The count of unique IPs returned in a successful test response, or -1 if   no responses were successful} \\
    more\_IPs                              & Discrete              & N                                  & Y                & NA                   & \makecell{True if there are more than 5 IP addresses in the test response (this occurs in   \textless 1\% of successful probes)} \\
    include\_IP\_i                         & Discrete              & N                                  & Y                & 5                    & \makecell{True if this IP slot is used} \\
    test\_response\_i\_IP\_match           & Discrete              & Y                                  & Y                & 5                    & \makecell{1 if the IP address matches at least one of those returned by the test   query response on the control resolver,\\0 if no match, and -1 if this IP   address slot is unfilled} \\
    test\_response\_i\_http\_match         & Discrete              & Y                                  & Y                & 5                    & \makecell{1 if the hash of the webpage matches at least one of those returned by   the test query response on the control\\resolver, 0 if no match, and -1 if   this IP address slot is unfilled} \\
    test\_response\_i\_cert\_match         & Discrete              & Y                                  & Y                & 5                    & \makecell{1 if the certificate hash matches that of at least one of the certificate   hashes from the test query response\\on the control resolver, 0 if no match,   and -1 if this IP address slot is unfilled} \\
    test\_response\_i\_asnum\_match        & Discrete              & Y                                  & Y                & 5                    & \makecell{1 if the AS number obtained from the MaxMind/Censys of the IP address   matches that of at least one IP address\\returned in the test query response   on the control resolver; 0 if no match; and -1 if this IP address slot is   unfilled} \\
    test\_response\_i\_asname\_match       & Discrete              & Y                                  & Y                & 5                    & \makecell{1 if the AS name obtained from the MaxMind/Censys of the IP address   matches that of at least one IP address\\returned in the test query response   on the control resolver; 0 if no match; and -1 if this IP address slot is   unfilled} \\
    test\_response\_ i\_match\_percentage  & Continuous            & N                                  & N                & 5                    & \makecell{The percentage of the five match elements that are true; -1 if this IP   address slot is unfilled} \\
    test\_response\_i\_asnum (*)           & Discrete              & Y                                  & N                & 5                    & \makecell{The AS number obtained from MaxMind/Censys for the IP address returned in   this response; -1 otherwise} \\
    average\_matchrate                     & Continuous            & N                                  & N                & NA                   & \makecell{The average match\_percentage among all the type-A records returned for   this probe} \\
    \bottomrule
    \end{tabular}}
    \caption{Description for the features extracted from Satellite that are used in training our machine learning models.}
\label{tab:satellite_features}
\end{table*}

\section{Additional results}
\label{appsec:additional_results}

\myparab{New DNS censorship signatures.} Tables~\ref{tab:ips} provides a list of
example new DNS injection signatures found across models that have been missed
by OONI and GFWatch.

\begin{table}[t]
	\centering
	\resizebox{.9\columnwidth}{!}{
		\begin{tabular}{llll}
			\toprule
			\textbf{Example IPs} & \textbf{\makecell{Missed by\\OONI}} & \textbf{\makecell{Missed by\\GFWatch}} & \textbf{\makecell{Missed by\\Both}} \\
			\midrule
            199.19.54.1 & Y & Y & Y \\
            199.19.56.1 & Y & Y & Y \\
            199.19.57.1 & Y & Y & Y \\
            203.98.7.65 & Y & N & N \\
            8.7.198.45 & Y & N & N \\
            59.24.3.173 & Y & N & N \\
            243.185.187.39 & Y & N & N \\
            4.36.66.178 & Y & N & N \\
            203.161.230.171 & Y & N & N \\
            157.240.18.18 & Y & N & N \\
            174.37.54.20 & Y & N & N \\
            74.86.3.208 & Y & N & N \\
            64.33.88.161 & Y & N & N \\
			\bottomrule
	    \end{tabular}
	}
	\caption{Example instances of injected forged IPs detected across models when using OONI data.}
	\label{tab:ips}
\end{table}

\myparab{Cross-country generalizability.} To show that the generalizability of
our models, i.e., models trained on data from one country will generalize to
another, we dropped all the country- and region-specific features such as ASN
from our data before retraining. As shown in
Table~\ref{tab:satelliteasndropping}, even without these features, both
supervised and unsupervised models perform well.
\begin{table}[!htbp]
	\centering
	\resizebox{\columnwidth}{!}{
		\begin{tabular}{lclllllll}
			\toprule
			Model & \makecell{ASN\\included} & TPR& FPR& TNR & FNR& Acc.&Prec.\\
			\midrule
			\makecell{Supervised\\(XGBoost)} & Y  & \textbf{99.4} & 0.9 &99.1&0.6&  99.1         & 95.4 \\
			\makecell{Supervised\\(XGBoost)} & N  & \textbf{99.4} &1.4&98.6&     0.6      & 98.7& 93.3\\
			\midrule
			\makecell{Unsupervised\\(OCSVM\_SGD)}      & Y              & \textbf{99.2} &19.3  &80.7&0.7& 83.8 &  50.1  \\
   			\makecell{Unsupervised\\(OCSVM\_SGD)}      & N & \textbf{99.1}    & 17.4&82.6&0.9& 85.3  &   52.7          \\
			\bottomrule
	\end{tabular}}

	\caption{Satellite model performance with and without including ASN in the features when trained,
 validated, and tested using GFWatch labels.}
	\label{tab:satelliteasndropping}
\end{table}

\myparab{Localized censorship records from OONI data.} Further investigation on
OONI's data in Table~\ref{tab:ooni_one_month} show many DNS anomalies are due to
probes conducted in autonomous systems (ASes) allocated with a small number of
IPs (e.g., AS17926, AS140308) that may experience localized filtering policies,
and may not represent China's national-level censorship as they are inconsistent.

\begin{table*}[!htbp]
	\centering
	\resizebox{\textwidth}{!}{
		\begin{tabular}{l|llllll}
			\toprule
			ASN & \makecell{Number of IPs} & \makecell{Domains Tested} & \makecell{Number of\\OONI DNS Blockings} & \makecell{Number of\\Inconsistent OONI\\DNS Blockings} & \makecell{Inconsistency Rate} & \makecell{Contribution to\\Total Inconsistent\\Measurements} \\
			\midrule
			AS134775 & 0.3M & 2 & 0 & 0 & 0.00\% & 0.00\% \\
			AS56040 & 2M & 1 & 0 & 0 & 0.00\% & 0.00\% \\
			AS56047 & 0.7M & 15 & 1 & 0 & 0.00\% & 0.00\% \\
			AS132525 & 0.4M & 2 & 1 & 0 & 0.00\% & 0.00\% \\
			AS4538 & 17M & 1 & 1 & 1 & 100.0\% & 0.18\% \\
			AS56044 & 1.0M & 2 & 1 & 0 & 0.00\% & 0.00\% \\
			AS17622 & 0.7M & 12 & 2 & 0 & 0.00\% & 0.00\% \\
			AS56048 & 1.2M & 11 & 5 & 1 & 20.00\% & 0.18\% \\
			AS4812 & 8.6M & 8 & 4 & 0 & 0.00\% & 0.00\% \\
			AS17816 & 4.0M & 30 & 4 & 0 & 0.00\% & 0.00\% \\
			AS4808 & 8.3M & 40 & 17 & 4 & 23.53\% & 0.71\% \\
			AS17926 & 0.7M & 2033 & 117 & 52 & 44.44\% & 9.22\% \\
			AS9808 & 36.4M & 460 & 134 & 2 & 1.49\% & 0.35\% \\
			AS134773 & 0.2M & 1289 & 392 & 15 & 3.83\% & 2.66\% \\
			AS17624 & 1.0M & 1421 & 429 & 23 & 5.36\% & 4.08\% \\
			AS4837 & 60M & 1640 & 449 & 22 & 4.90\% & 3.90\% \\
			AS56046 & 3.0M & 1664 & 480 & 34 & 7.08\% & 6.03\% \\
			AS140308 & 0.2M & 1847 & 817 & 332 & 40.64\% & 58.87\% \\
			AS4134 & 110.7M & 1970 & 564 & 79 & 14.01\% & 14.01\% \\
			\midrule
			Total & 255.6M & 12448 & 3418 & 564 & 16.50\% & 100.00\% \\
			\bottomrule
		\end{tabular}
	}
	\caption{A comparison of inconsistency rates and contributions to total inconsistent measurements between OONI and GFWatch.}
	\label{tab:ooni_one_month}
\end{table*}

\myparab{Feature weights for OONI and Censored Planet data.}
Tables~\ref{tab:ooni_fi} and~\ref{tab:satellite_fi} contain the feature
importance of the optimal models chosen in this study. For models including
OCSVM\_SGD and XGBoost, the SHAP values \cite{lundberg2017unified} are used to
measure the feature contributions to the prediction results. For Isolation
Forest models, the feature importance of features are computed by measuring
their average frequencies of utilization by the different trees. We provide the
top 10 features for all models because features beyond this range show
negligible importance.

\begin{table*}[!htbp]
	\centering
	\resizebox{\textwidth}{!}{
		\begin{tabular}{*{5}{l}}
			\toprule
			\textbf{Rank} & \textbf{IF (OONI Labels)} & \textbf{IF (GFWatch Labels)} & \textbf{XGBoost (OONI Labels)} & \textbf{XGBoost (GFWatch Labels)} \\
			\midrule
			1  & test\_keys\_asn (0.131)  & test\_keys\_asn (0.131)  & test\_keys\_asn (0.142)  & resolver\_asn (0.138)  \\
			2  & resolver\_network\_name (0.128)  & resolver\_network\_name (0.128)  & resolver\_asn (0.135)  & test\_keys\_asn (0.133)  \\
			3  & resolver\_asn (0.121)  & resolver\_asn (0.121)  & test\_keys\_as\_org\_name (0.132)  & test\_keys\_as\_org\_name (0.128)  \\
			4  & probe\_asn (0.121)  & probe\_asn (0.121)  & probe\_asn (0.122)  & probe\_network\_name (0.122)  \\
			5  & test\_keys\_as\_org\_name (0.120)  & test\_keys\_as\_org\_name (0.120)  & probe\_network\_name (0.117)  & probe\_asn (0.118)  \\
			6  & probe\_network\_name (0.118)  & probe\_network\_name (0.118)  & resolver\_network\_name (0.109)  & resolver\_network\_name (0.092)  \\
			7  & http\_experiment\_failure (0.083)  & http\_experiment\_failure (0.083)  & http\_experiment\_failure (0.062)  & http\_experiment\_failure (0.084)  \\
			8  & status\_code\_match (0.029)  & status\_code\_match (0.029)  & status\_code\_match (0.046)  & status\_code\_match (0.041)  \\
			9  & title\_match (0.028)  & title\_match (0.028)  & body\_length\_match (0.023)  & body\_length\_match (0.026)  \\
			10 & body\_length\_match (0.027)  & body\_length\_match (0.027)  & headers\_match (0.022)  & headers\_match (0.025)  \\
			\bottomrule
	\end{tabular}}
	\caption{Top 10 important features for Isolation Forests (IF) and XGBoost classifiers when using OONI data.}
	\label{tab:ooni_fi}
\end{table*}

\begin{table*}[!htbp]
	\centering
	\resizebox{.6\textwidth}{!}{%
		\begin{tabular}{lll}
			\toprule
			\textbf{Rank} & \textbf{OCSVM\_SGD (GFWatch Labels)} & \textbf{XGBoost (GFWatch Labels)} \\
			\midrule
			1 & test\_noresponse\_i\_rcode (0.102) & test\_response\_i\_asnum (0.912) \\
			2 & untagged\_response (0.096) & test\_response\_i\_IP\_match (0.010) \\
			3 & test\_response\_i\_match\_pct (0.081) & test\_response\_i\_asnum\_match (0.009) \\
			4 & test\_noresponse\_i\_has\_type\_a (0.080) & test\_noresponse\_i\_rcode (0.009) \\
			5 & test\_response\_i\_IP\_match (0.074) & test\_response\_i\_asname\_match (0.009) \\
			6 & test\_response\_i\_asnum\_match (0.069) & test\_response\_i\_cert\_match (0.009) \\
			7 & test\_response\_i\_asname\_match (0.068) & test\_response\_i\_http\_match (0.009) \\
			8 & test\_response\_i\_http\_match (0.056) & test\_noresponse\_i\_has\_type\_a (0.006) \\
			9 & test\_response\_i\_cert\_match (0.056) & include\_IP\_i (0.006) \\
			10 & test\_query\_successful (0.054) & test\_response\_i\_match\_pct (0.003) \\
			\bottomrule
		\end{tabular}%
	}
	\caption{Top 10 feature importance for OCSVM\_SGD and XGBoost classifiers when using Satellite data.}
	\label{tab:satellite_fi}
\end{table*}

\end{sloppypar}
\end{document}